\pdfoutput=1

\documentclass[11pt]{article}

\usepackage{emnlp2021}
\usepackage{times}
\usepackage{latexsym}

\usepackage[T1]{fontenc}

\usepackage[utf8]{inputenc}

\usepackage{microtype}

\usepackage{graphicx}
\usepackage{multirow}
\usepackage{arydshln}
\usepackage{lscape}
\usepackage{tipa}
\usepackage{transparent}
\usepackage{svg}

\def\Snospace~{\S{}}
\usepackage{hyperref}
\usepackage{cleveref}
\usepackage{changepage}

\definecolor{airforceblue}{rgb}{0.36, 0.54, 0.66}

\newcommand{\ia}{\textit{inter alia}} 
\newcommand{\ie}{\textit{i.e.}} 
\newcommand{\eg}{\textit{e.g.}} 
\newcommand{\predicate}[1]{\texttt{#1}} 
\newcommand{\pbrole}[1]{\texttt{#1}} 
\newcommand{\role}[1]{\textsc{#1}} 

\title{Asking It All: Generating Contextualized Questions\\ for any Semantic Role}

\author{%
Valentina Pyatkin$\thanks{~~Equal contribution}$ $^{\,,1}$%
~\;~\;~ Paul Roit$^{*,1}$%
~\;~\;~ Julian Michael$^{2}$\\
{\bf~\;~\;~ Reut Tsarfaty$^{1,3}$}
{\bf~\;~\;~ Yoav Goldberg$^{1,3}$}
{\bf~\;~\;~ Ido Dagan$^{1}$}\\
  $^{1}$Computer Science Department, Bar Ilan University\\%
  ~\;~\;~\;~$^{2}$Paul G. Allen School of Computer Science \& Engineering, University of Washington\\%
  ~\;~\;~\;~$^{3}$Allen Institute for Artificial Intelligence\\
  \texttt{\footnotesize{\{valpyatkin,plroit,yoav.goldberg\}@gmail.com}}\\\texttt{\footnotesize{julianjm@cs.washington.edu}}%
  ,~\;~ \texttt{\footnotesize{reut.tsarfaty@biu.ac.il}},~\;~ \texttt{\footnotesize{dagan@cs.biu.ac.il}}
 }

\begin{document}
\maketitle

\begin{abstract}
Asking questions about a situation is an inherent step towards understanding it. 
To this end, we introduce the task of role question generation, which, given a predicate mention and a passage, requires producing a set of questions asking about all possible semantic roles of the predicate. We develop a two-stage model for this task, which first produces a context-independent question prototype for each role and then revises it to be contextually appropriate for the passage.
Unlike most existing approaches to question generation, our approach does not require conditioning on existing answers in the text.
Instead, we condition on the type of information to inquire about, regardless of whether the answer appears explicitly in the text, could be inferred from it, or should be sought elsewhere.
Our evaluation demonstrates that we generate diverse and well-formed questions for a large, broad-coverage ontology of predicates and roles.
\end{abstract}

\section{Introduction}

Soliciting information by asking questions is an essential communicative ability, and natural language question-answer (QA) pairs provide a flexible format for representing and querying the information expressed in a text. This flexibility has led to applications in a wide range of tasks from reading comprehension \citep{rajpurkar-etal-2016-squad} to information seeking dialogues \citep{qi-etal-2020-stay}.

\begin{figure}[t]
\resizebox{\columnwidth}{!}{%
\begin{tabular}{ll}
\multicolumn{2}{p{3in}}{ \textcolor{olive}{The plane} took off in \textcolor{teal}{Los Angeles}. \textcolor{blue}{The tourists} will \textbf{arrive} \textcolor{brown}{in Mexico} \textcolor{cyan}{at noon.}} \\ \hline
entity in motion        &\textcolor{blue}{Who will arrive in Mexico?}  \\ 
end point   & \textcolor{brown}{Where will the tourists arrive?}      \\ 
start point  & \textcolor{teal}{Where will the tourists arrive from?} \\
manner   & \textcolor{olive}{How will the tourists arrive?}\\ 
cause   & \textcolor{gray}{Why will the tourists arrive?}\\ 
temporal   & \textcolor{cyan}{When will the tourists arrive?}    \\ 
\end{tabular}}
        \caption{Example role questions for ``arrive''. Some questions are for explicit arguments (\textit{entity in motion},\textit{end point}, \textit{temporal}), some for implicit (\textit{start point}, \textit{manner}) ones, and some for arguments that do not appear at all (\textit{cause}).}
    \label{fig1}
\end{figure}

Automatically generating questions can potentially serve as an essential building block for such applications. Previous work in question generation has either required human-curated templates \citep{levy-etal-2017-zero,du-cardie-2020-event}, limiting coverage and question fluency, or generated questions for answers already identified in the text \citep{heilman-smith-2010-good,dhole2020syn}.
Open-ended generation of information-seeking questions, where the asker poses a question inquiring about a certain type of information, remains a significant challenge.

In this work, we propose to resolve these difficulties by generating \textit{role questions}.
In particular, for any predicate expressed in a text and semantic role associated with that predicate, we show how to generate a contextually-appropriate question
whose answer---if present---corresponds to an argument of the predicate bearing the desired role. Some examples are shown in \autoref{fig1}.
Since the set of possible questions is scoped by the relations in the underlying ontology, this gives us the ability to \textit{ask it all}: pose information-seeking questions that exhaustively cover a broad set of semantic relations that may be of interest to downstream applications.

Concretely, we generate questions derived from QA-SRL \citep{he-etal-2015-question} for the semantic role ontology in PropBank \citep{palmer-etal-2005-proposition} using a two-stage approach.
In the first stage (\S\ref{sec:generating_protos}), we leverage corpus-wide statistics to compile an ontology of simplified, context-independent \textit{prototype questions} for each PropBank role. 
In the second stage (\S\ref{sec:contextualizing}), we contextualize the question using a learned translation from prototypes to their contextualized counterparts, conditioned on a sentence. To that end we present a new resource of \textit{frame-aligned QA-SRL questions} which are grounded in their source context. 
This setup decouples posing a question that captures a semantic role from fitting that question to the specific context of the passage.
As we show in \S\ref{analysis_evaluation}, the result is a system which generates questions that are varied, grammatical, and contextually appropriate for the passage, and which correspond well to their underlying role.\footnote{Our code and resources can be found here: \url{https://github.com/ValentinaPy/RoleQGeneration}}

The ability to exhaustively enumerate a set of questions corresponding to a known, broad-coverage underlying ontology of relations allows for a comprehensive, interpretable, and flexible way of representing and manipulating the information that is contained in---or missing from---natural language text.
In this way, our work takes an essential step towards combining the advantages of formal ontologies and QA pairs for broad-coverage natural language understanding.

\section{Background}

\begin{table}
\centering
\resizebox{\columnwidth}{!}{%
\begin{tabular}{llllllll}
WH    & AUX   & SBJ       & VERB   & OBJ       & PREP & MISC    & ? \\
Who   & might &           & bring  & something & to   & someone & ? \\
Where & would & someone   & arrive &           & at   &         & ? \\
What  & was   & something & sold   &           & for  &         & ?
\end{tabular}%
}
\caption{QA-SRL slot format. WH and VERB are mandatory, and AUX and VERB encode tense, modality, negation, and active/passive voice.
}
\label{tab:qasrl}
\end{table}

\paragraph{Question Generation}
Automatic question generation has been employed for use cases such as
constructing educational materials \citep{mitkov-ha-2003-computer},
clarifying user intents \citep{Aliannejadi2019AskingCQ},
and eliciting labels of semantic relations in text \citep{fitzgerald-etal-2018-large, klein-etal-2020-qanom, pyatkin-etal-2020-qadiscourse}.
Methods include
transforming syntactic trees \citep{heilman-smith-2010-good,dhole2020syn}
and SRL parses \citep{mazidi2014linguistic,flor2018semantic},
as well as training seq2seq models conditioned on the question's answer \citep{fitzgerald-etal-2018-large} or a text passage containing the answer \citep{du2017learning}.
By and large, these approaches are built to generate questions whose answers are already identifiable in a passage of text.

However, question generation has the further potential to seek \textit{new} information, which requires asking questions whose answers may only be implicit, inferred, or even absent from the text.
Doing so requires prior specification of the kind and scope of information to be sought. As a result, previous work has found ways to align existing relation ontologies with questions, either through human curation \citep{levy-etal-2017-zero} --- which limits the approach to very small ontologies --- or with a small set of fixed question templates \citep{du-cardie-2020-event} --- which relies heavily on glosses provided in the ontology, sometimes producing stilted or ungrammatical questions.
In this work, we generate natural-sounding information-seeking questions for a broad-coverage ontology of relations such as that in PropBank \citep{bonial2014propbank}.

\paragraph{QA-SRL}\label{qasrl_formalism}
Integral to our approach is QA-SRL \cite{he-etal-2015-question}, a representation based on question-answer pairs which was shown by \citet{roit-etal-2020-controlled} and \citet{klein-etal-2020-qanom} to capture the vast majority of arguments and modifiers in PropBank and NomBank \citep{palmer-etal-2005-proposition,Meyers2004TheNP}.
Instead of using a pre-defined role lexicon, QA-SRL labels semantic roles with questions drawn from a 7-slot template, whose answers denote the argument bearing the role (see \autoref{tab:qasrl} for examples).
Unlike in PropBank, QA-SRL argument spans may appear outside of syntactic argument positions, capturing \textit{implicit} semantic relations \cite{gerber2010beyond, Ruppenhofer2010SemEval2010T1}.

QA-SRL is useful to us because of its close correspondence to semantic roles and its carefully restricted slot-based format: it allows us to easily transform questions into context-independent \textit{prototypes} which we can align to the ontology, by removing tense, negation, and other information immaterial to the semantic role (\autoref{sec:generating_protos}).
It also allows us to produce \textit{contextualized} questions which sound natural in the context of a passage, by automatically aligning the syntactic structure of different questions for the same predicate (\autoref{sec:contextualizing}).

\begin{figure*}[!htb]
\includegraphics[width=\textwidth]{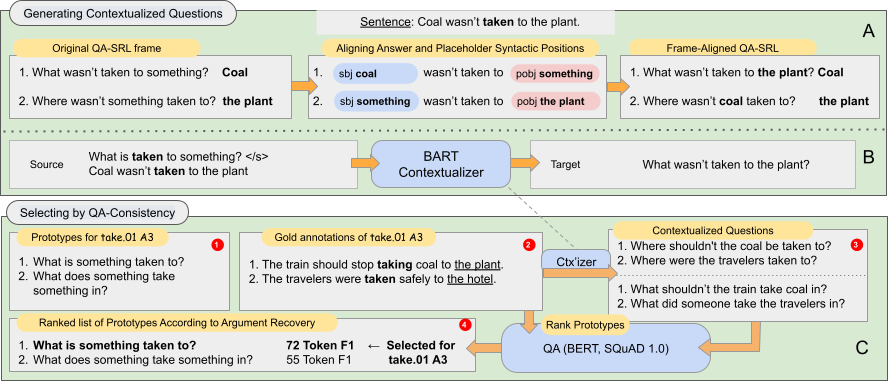}
\caption{\textbf{A: Construction of Frame-Aligned QA-SRL}
(\autoref{sec:contextualizing}) using syntactic information inferred by the autocomplete NFA from \citet{fitzgerald-etal-2018-large}.
\textbf{B: Contextualizing questions.} The contextualizer takes in a prototype question and a context, and outputs a Frame-Aligned QA-SRL question. Note how Frame-Aligned QA-SRL questions preserve the role expressed in their respective prototypes.
\textbf{C: Selecting prototype questions} according to their ability to recover explicit arguments (see \autoref{sec:generating_protos}).
We test each prototype (1) against a sample of arguments for each role (2). Contextualized versions of each prototype question for each sampled sentence (3) are fed (with the sentences) to a QA model, and the predicted answer is evaluated against the gold argument. The highest ranking prototype (4) is selected for that role. }
\label{fig:overview}
\end{figure*}

\section{Task Definition} \label{task_def}

Our task is defined with respect to an ontology of semantic roles such as PropBank. Given a passage of text with a marked predicate and a chosen role of that predicate, we aim to generate a naturally-phrased question which captures that semantic relation. For example, consider \autoref{fig1}. For the predicate \predicate{arrive.01} and role \role{A0} (defined in PropBank as \textit{entity in motion}), we want to generate the question \textit{Who will arrive in Mexico?}.

In this task, the lack of an answer in the text should not preclude the system from generating a question that pertains to a role. Rather, the question should be such that \textit{if} an answer exists, it denotes an argument bearing that role.

\section{Method}

Generating role questions requires mapping from an input sentence, a predicate, and role to a natural language question, regardless of whether the question is answerable in the text.
However, we don't have training data of this nature: existing question-answer driven approaches to semantic annotation \citep{fitzgerald-etal-2018-large} only elicit \textit{answerable} questions from annotators, and existing resources with unanswerable questions (\citet{rajpurkar2018know}, \ia) do not systematically align to semantic roles in any clear way.
Indeed, we will show in \autoref{sec:main-eval} that training a seq2seq model to directly generate role questions from question-answer pairs annotated with explicit semantic roles yields poor results for implicit or missing arguments, because the model overfits towards only asking answerable questions.

Instead, we adopt a \textbf{two-stage approach}: first, we leverage corpus-wide statistics to map each role in an ontology to a context-independent \textbf{prototype question} which provides the high-level syntactic structure of any question for that role.
For example, \predicate{bring.01}'s \textit{destination} argument \role{A2} may receive the prototype question \textit{where does something bring something?}
Second, we employ a \textbf{question contextualization} step which aligns the tense, negation/modality, and entities in the question with those in the particular source text, e.g., \textit{where did the pilot bring the plane?}.
The result is a question about the source text which is amenable to traditional QA systems and whose semantically correct answer should relate to the predicate according to the given role.
\autoref{tab:inference} shows example prototype and contextualized questions for the predicate \predicate{change.01}.\footnote{In this work, we target both verbal predicates (\eg, \textit{eat}, \textit{give}, \textit{bring}) and deverbal nouns (such as \textit{sale}, \textit{presentation}, and \textit{loss}), using the frame index associated with the OntoNotes corpus \citep{Weischedel2017OntoNotesA}. In addition to the core roles for each predicate, we generate questions for the adjunct roles \textsc{Locative, Temporal, Manner, Causal, Extent,} and \textsc{Goal}.}

\begin{figure*}[t]
\centering
\resizebox{\textwidth}{!}{%
\begin{tabular}{lll}
causer of transformation         &  \textcolor{brown}{Who changes something?} & \textcolor{blue}{Who might have changed their minds? } \\
thing changing &  \textcolor{brown}{What is changed ?}   & \textcolor{blue}{What might have been changed?}   \\
end state           &  \textcolor{brown}{What is something changed to?} & \textcolor{blue}{What might their minds be changed to?} \\
start state &  \textcolor{brown}{What is changed into something? }      &\textcolor{blue}{What might have been changed into something?} 
\end{tabular}%
}
\caption{Inference for change.01 given the sentence: \textit{"The only thing that might've changed their minds this quickly I think is money"}. For each role (left) we look-up a context-independent question (center) and apply a contextualizing transformation to produce a sound question (right).}
\label{tab:inference}
\end{figure*}

\subsection{Generating Prototype Questions}\label{sec:generating_protos}
In our first step, we introduce \textit{prototype questions} to serve as intermediate, \textit{context-independent} representations of each role in the ontology. We obtain these prototypes by automatically producing a large dataset of QA-SRL questions jointly labelled with PropBank roles. Each question is then stripped of features which are irrelevant to the underlying semantic role, and the resulting prototypes are aggregated for each role. Finally, we choose the prototype which
allows a question answering model to most accurately recover the arguments corresponding to that role.
The end result is the assignment of a single prototype question to each semantic role in the ontology, to be fed into the second step (\autoref{sec:contextualizing}).

\paragraph{Joint PropBank and QA-SRL}
We begin by aligning PropBank roles with QA-SRL questions.
We do this in two ways:
First, we run the SRL parser\footnote{We use AllenNLP's \cite{Gardner2018AllenNLPAD} public model for verbal SRL, and train our own parser on the original NomBank data.} of \citet{Shi2019SimpleBM} on the source sentences of the QA-SRL Bank 2.0 \citep{fitzgerald-etal-2018-large} and QANom \citep{klein-etal-2020-qanom}.
To be aligned, the answer must have significant ($\geq$0.4 intersection-over-union) overlap with the predicted SRL argument and the question must target the same predicate.
Second, we run the QA-SRL question generator from \citet{fitzgerald-etal-2018-large} on the gold predicates and arguments in the OntoNotes training set to produce QA-SRL questions aligned to each argument.
Altogether, this produces a total of 543K instances of jointly-labeled PropBank arguments and QA-SRL questions.
We manually check 200 sampled instances of both verbal and nominal aligned QA pairs, split equally, and find that 93\% had no errors.

\paragraph{Aggregating Prototype Candidates}\label{par:qasrl_protos}
For each role, we enumerate the full set of QA-SRL questions appearing for that role in our jointly labeled data.\footnote{To increase coverage, we share prototypes for the same role label between different senses of the same predicate.}
We post-process the QA-SRL questions into what we call \textit{question prototypes} using the same process as \citet{michael-zettlemoyer-2021-inducing}: we remove negations, replace animate pronouns (\textit{who}/\textit{someone}) with inanimate ones (\textit{what}/\textit{something}), and convert all questions to the simple present tense (which can be easily done in QA-SRL's slot-based format).
This eliminates aspects of a question which are specific of a particular text, while retaining aspects that are potentially indicative of the underlying semantic role, such as the question's syntactic structure, active/passive distinction, prepositions, and \textit{wh}-word (\eg, \textit{when/where}).
For example, the questions \textit{What will be fixed?} and \textit{What won't be fixed?} have conflicting meanings, but both target the same semantic role (the \role{Theme} of \predicate{fix}) and they receive the same prototype (\textit{What is fixed?}).
Full details of this method are described in \autoref{sec:prototype_conversion}.

\paragraph{Selecting by QA Consistency}
\label{cycle-consistency}
Of the prototype candidates for each role, we want to choose the one with the right specificity:
for example, \textit{where does something win?} is a more appropriate prototype for an \pbrole{AM-LOC} modifier role than than \textit{what does someone win at?}, even if the latter appears for that role.
This is because \textit{what does someone win at?} is at once too specific (inappropriate for locations better described with \textit{in}, \eg, \textit{in Texas}) and too general (also potentially applying to \predicate{win.01}'s \pbrole{A2} role, the contest being won).

To choose the right prototype, we run a consistency check using an off-the-shelf QA model (see \autoref{fig:overview}, Bottom).
We sample a set of gold arguments\footnote{For core roles we sample 50 argument instances, and for adjunct roles we take 100 but select samples from any predicate sense.} for the role from OntoNotes \cite{Weischedel2017OntoNotesA} and instantiate each prototype for each sampled predicate using the question contextualizer described in the next section (\autoref{sec:contextualizing}).
We then select the prototype for which a BERT-based QA model \cite{devlin-etal-2019-bert} trained on SQuAD 1.0 \citep{rajpurkar-etal-2016-squad} achieves the highest token-wise F1 in recovering the gold argument from the contextualized question.
\newcommand{\colorx}[1]{{\color{red}\textbf{#1}}}
\newcommand{\colory}[1]{{\color{blue}\textbf{#1}}}

\begin{figure}
    \begin{adjustwidth}{.5em}{.5em}
        \textit{Air molecules move a lot and \textbf{bump} into things.}
    \\[0.3em]
    \textbf{QA-SRL:}
    \begin{adjustwidth}{1em}{1em}
        What bumps into something?
        \begin{adjustwidth}{1em}{1em}
        $\hookrightarrow$ \colory{Air molecules}
        \end{adjustwidth}
        What does something bump into?
        \begin{adjustwidth}{1em}{1em}
        $\hookrightarrow$ \colorx{things}
        \end{adjustwidth}
    \end{adjustwidth}
    \textbf{Syntactic Alignment:}
    \begin{adjustwidth}{1em}{1em}
    \begin{tabular}{@{}ll}
        tense:  & \textit{present} \\
        SUBJ:   & \colory{Air molecules} \\
        OBJ:    & $\emptyset$ \\
        PP:     & into \colorx{things}
    \end{tabular}
    \end{adjustwidth}
    \textbf{Frame-Aligned QA-SRL:}
    \begin{adjustwidth}{1em}{1em}
    What bumps into \colorx{things}?
    \begin{adjustwidth}{1em}{1em}
    $\hookrightarrow$ \colory{Air molecules}
    \end{adjustwidth}
    What do \colory{air molecules} bump into?
    \begin{adjustwidth}{1em}{1em}
    $\hookrightarrow$ \colorx{things}
    \end{adjustwidth}
    \end{adjustwidth}
    \end{adjustwidth}
    \caption{Example QA-SRL contextualization. The autocomplete NFA from \citet{fitzgerald-etal-2018-large} keeps track of the syntactic position of the gap produced by wh-movement, which allows us to substitute the answer of one question in place of the placeholder pronouns in another. We also use simple heuristics to correct capitalization and a masked language model to correct verb agreement.}
    \label{fig:aligned-qasrl-example}
\end{figure}
\subsection{Generating Contextualized Questions}\label{sec:contextualizing}
For our second stage, we introduce a \textit{question contextualizer} model which takes in a prototype question and passage, and outputs a contextualized version of the question, designed to sound as natural as possible and match the semantics of the passage (see \autoref{fig:overview}, Top).
In particular, our model adjusts the tense, negation, and modality of the question and fills in the placeholder pronouns with their corresponding mentions in the surrounding context.

To train the contextualizer, we automatically construct a new resource --- the Frame-Aligned QA-SRL Bank --- which pairs QA-SRL question prototypes
with
their contextualized forms (see \autoref{fig:aligned-qasrl-example} for an example).
The latter are constructed on the basis of syntactic alignments between different questions about the same predicate instance. 
We then train a BART model \citep{lewis-etal-2020-bart} on this data to directly perform contextualization.

\paragraph{Recovering Syntax from QA-SRL}
The first step to constructing contextualized QA-SRL questions is identifying the questions' underlying syntactic structure, which can be used to align each question's answers with the placeholder positions of other questions in the same frame.
We use the autocomplete NFA from \citet{fitzgerald-etal-2018-large}, which constructs a simple syntactic representation of QA-SRL questions with three grammatical functions: subject, object, a third argument which may be an object, prepositional phrase, or complement.
This recovers the declarative form of a QA-SRL question: for example, \textit{What bumps into something?} corresponds to the declarative clause \textit{
Something
bumps into something%
}, and asks about the argument in the subject position.
Questions with adverbial \textit{wh}-words, like \textit{When did something bump into something?}, are mapped to a declarative clause without the adverbial (\ie, \textit{Something bumped into something}).

In some cases, the syntax is ambiguous: for example, the question \textit{What does something bump into?} may correspond to either \textit{Something bumps something into} or \textit{Something bumps into something}.
To resolve ambiguities, we choose the interpretation of each question which is shared with the most other questions in the same frame.
To handle ties, we fall back to heuristics listed in \autoref{app:syntactic-ambiguity}.

\paragraph{Aligning Answers with Placeholders} \label{sec:juliandata}
Where multiple questions share their underlying syntax,
we replace the placeholder pronouns (e.g., \textit{something} or \textit{someone}) in each with the answers corresponding to their syntactic position (see \autoref{fig:overview}, Top, and \autoref{fig:aligned-qasrl-example}).
To increase coverage of placeholder pronouns, we extend the correspondences to hold between slightly different syntactic structures, e.g., the passive subject and transitive object.
Finally, we correct capitalization with simple heuristics and fix subject-verb agreement using a masked language model.
Full details are in \autoref{app:aligning-answers}.
Altogether, this method fills in 91.8\% of the placeholders in the QA-SRL Bank 2.0 and QANom.

\paragraph{Model}
We train a BART model \cite{lewis-etal-2020-bart} taking a passage with a marked predicate and a prototype question as input, where the output is the contextualized version of the same question provided in the new Frame-Aligned QA-SRL Bank. This setup ensures that the basic structure indicative of the semantic role is left unchanged, since both the source and target questions are derived from the same original using role preserving transformations. As a result, the model learns to preserve the underlying semantic role, and to retain the tense, modality, negation and animacy information of the original question. Full training details are in \autoref{app:training-contextualizer}.

Equipped with a question contextualizer, we can perform the full task: given a predicate in context and desired role, we retrieve the question prototype for the role (\autoref{sec:generating_protos}), pair it with the context, and run it through the contextualizer to produce the final role question.

\section{Evaluation}\label{analysis_evaluation} 

\begin{table*}[t]
\resizebox{\textwidth}{!}{%
\begin{tabular}{ll|llll|llll|ll|ll|l}
                       &          & \multicolumn{4}{l}{Role Accuracy}       & \multicolumn{4}{l}{QA Accuracy} & \multicolumn{2}{l}{Grammaticality} & \multicolumn{2}{l}{Adequacy} &       \\
                       &          & All     & Expl.    & Impl.       & None       & All   & Expl.  & Impl.  & None  & All             & GRM+RC           & All          & ADQ+RC        & Freq. \\ \hline
Onto* & RoleQ    & 0.72    & 0.81     & \multicolumn{2}{c|}{0.64} & -     & 0.75   & -      & -     & 4.25            & 60\%              & 4.05         & 56\%           & 1210  \\ \hdashline \hdashline
\multirow{2}{*}{Onto}  & RoleQ    & 0.78    & 0.93     & \multicolumn{2}{c|}{0.64} & -     & 0.81   & -      & -     & 4.22            & 61\%              & 4.07         & 58\%           & 289   \\
                       & e2e      & 0.51    & 0.88     & \multicolumn{2}{c|}{0.25} & -     & 0.85   & -      & -     & 4.34            & 49\%              & 4.20         & 46\%           & 289   \\ \hdashline
\multirow{2}{*}{G\&C}  & RoleQ    & 0.76    & 0.82     & 0.80        & 0.63       & 0.50  & 0.36   & 0.58   & 0.57  & 4.45            & 69\%               & 4.20         & 58\%            & 120   \\
                       & e2e      & 0.54    & 0.66     & 0.57        & 0.32       & 0.37  & 0.33   & 0.43   & 0.32  & 4.74            & 50\%               & 4.53         & 51\%            & 120   \\ \hdashline
\multirow{2}{*}{ON5V}  & RoleQ    & 0.85    & 0.83     & 0.92        & 0.79       & 0.50  & 0.59   & 0.40   & 0.39  & 4.57            & 78\%              & 4.24         & 69\%           & 148   \\
                       & e2e      & 0.63    & 0.86     & 0.44        & 0.31       & 0.46  & 0.74   & 0.19   & 0.09  & 4.81            & 61\%               & 4.69         & 59\%            & 148  
\end{tabular}}
        \caption{Analysis of our questions on multiple splits by SRL argument types (Explicit, Implicit, and None / not present) over a sample of predicates in OntoNotes, G\&C and ON5V. GRM+RC and ADQ+RC are the percentage of questions that were rated with GRM (resp. ADQ) $\geq 4$ and also aligned to the correct description.
        RoleQ is our model, and e2e is the baseline.
        Onto* is the full OntoNotes set of 400 predicates. } 
    \label{tag:intrinsic}
\end{table*}

To assess our system, we perform an intrinsic evaluation against a seq2seq baseline (\autoref{sec:main-eval}) as well as comparisons to existing question generation systems (\autoref{sec:comparisons}).

\subsection{Metrics}
For our evaluations, we measure three properties of role questions.
First, \textbf{grammaticality}: is the question well-formed and grammatical?
Second, \textbf{adequacy}: does the question make sense in context? For this property, a question's presuppositions must be satisfied; for example, the question \textit{Who will bring the plane back to base?} is only adequate in a context where a plane is going to be brought to base. (Note that the answer does not necessarily have to be expressed.) 
Third, we measure \textbf{role correspondence}: does the question correspond to the correct semantic role?

For all of these measures, we source our data from existing SRL datasets and use human evaluation by a curated set of trusted workers on Amazon Mechanical Turk.\footnote{Screenshots of the task interfaces are in \autoref{app:interfaces}.}
Automated metrics like \textsc{Bleu} or \textsc{Rouge} are not appropriate for our case because our questions' meanings can be highly dependent on minor lexical choices (such as with prepositions) and because we lack gold references (particularly for questions without answers present).

We assess grammaticality and adequacy on a 5-point Likert scale, as previous work uses for similar measures \citep{elsahar2018zero,dhole2020syn}.
We measure role correspondence with two metrics: \textit{role accuracy}, which asks annotators to assign the question a semantic role based on PropBank role glosses, and \textit{question answering accuracy}, which compares annotators' answers to the question against the gold SRL argument (or the absence of such an argument).\footnote{Where an argument is present, we consider an annotator's answer correct if it includes its syntactic head.}

\subsection{Main Evaluation}
\label{sec:main-eval}

\paragraph{Data}
We evaluate our system on a random sample of 400 predicate instances (1210 questions) from Ontonotes 5.0 \citep{Weischedel2017OntoNotesA} and 120 predicate instances (268 questions) from two small implicit SRL datasets:  \citet[G\&C]{gerber2010beyond} and \citet[ON5V]{moor2013predicate}.
We generate questions for all core roles in each instance.
For comparison to the baseline, we use all predicates from the implicit SRL datasets and a subsample of 100 from OntoNotes. We also evaluate questions for 5 modifier roles\footnote{\textsc{Locative, Temporal, Manner, Causal, Extent,} and \textsc{Goal}} on 100 OntoNotes instances.

\paragraph{End-to-End Baseline}
As a baseline, we use a BART-based seq2seq model trained to directly generate role questions from a text passage with a marked target predicate, PropBank role label, and role description; for example: 
\texttt{Some geologists \textlangle{}p\textrangle{} study \textlangle{}/p\textrangle{} the Moon . \textlangle{}/s\textrangle{} student A0 study.01}, where \textit{student} is the gloss provided in PropBank for the \pbrole{A0} role of \predicate{study.01}. 
To train the model, we use the contextualized questions in the Frame-Aligned QA-SRL dataset (\autoref{sec:contextualizing}) as outputs while providing their aligned predicted PropBank roles (as detailed in \autoref{sec:generating_protos}) in the input.

\paragraph{Main Results}
Results are in \autoref{tag:intrinsic}.
We stratify on whether the argument matching the role was \textit{explicit} (syntactically related to the predicate), \textit{implicit} (present in the text, but not a syntactic argument), or \textit{None} (missing from the text entirely).
We combine \textit{implicit} and \textit{None} for OntoNotes since it only marks explicit arguments.

\paragraph{Role Correspondence}
\label{RC}
On role accuracy and QA accuracy, our model showed its strongest performance on explicit arguments, as well as implicit arguments in the G\&C/ON5V datasets.
It significantly outperforms the baseline on these metrics, with a 26 point gain in role accuracy and 11 point gain in QA accuracy on average.
The difference is much greater for implicit and missing arguments, with a 38 point gain for RC and a 23 point gain for QA, showing how our approach excels at producing questions with the correct semantics for these arguments, despite their unavailability during training.
Our results complement previous work \citep{moryossef-etal-2019-step} showing the utility of micro-planning with an intermediate representation to handle cases with little or no training data (\eg, implicit arguments).

\begin{table*}[t]
\small
\resizebox{\textwidth}{!}{%
\begin{tabular}{ll}
\multicolumn{2}{p{6.15in}}{[...] Jordan's King Abdullah II \textbf{pardoned} (\textsc{Justice.Pardon}/pardon.01) \textit{the former legislator known for her harsh criticism of the state} (\textsc{Defendant}/A1) .}\\
\textbf{EEQ} Who is the defendant? & \textbf{RoleQ} Who did Jordan's King Abdullah II pardon? \\ \hline
\multicolumn{2}{p{6.15in}}{[...] gun crime incidents are averaging about 29 a day in England and Wales, more than twice the level of when \textit{the Labour Government}(\textsc{Entity}/A1) \textbf{came} (\textsc{Personnel.Elect}/come.01) to power in 1997.} \\
\textbf{EEQ} Who voted? & \textbf{RoleQ} What came?\\ \hline
\multicolumn{2}{p{6.15in}}{About 160 workers at a factory that \textbf{made} \textit{paper} (A1) for the Kent filters were exposed to asbestos in the 1950s.}      \\       \textbf{Syn-QG}    What materials did a factory produce ?  & \textbf{RoleQ} What did 160 workers make? \\ \hline
\multicolumn{2}{p{6.15in}}{But you \textbf{believe} \textit{the fact that the U.S. Supreme Court just decided to hear this case is a partial victory for both Bush and Gore.} (A1)}\\
\textbf{Syn-QG} What do you believe in ? & \textbf{RoleQ} What is being believed? \\                                                                                                                                         
\end{tabular}}
        \caption{Examples of our role questions, Event Extraction Questions \cite{du-cardie-2020-event}, and Syn-QG  questions \cite{dhole2020syn}}
    \label{tab:exqcomparison}
\end{table*}

\paragraph{Grammaticality \& Adequacy}
Average grammaticality and adequacy scores are shown in \autoref{tag:intrinsic}.
In addition, for each measure we record the percentage of questions that both got a score $\geq$4 and were assigned to the correct role.
Our questions were deemed grammatical and adequate overall, with average scores above 4, but the baseline scored even better on all datasets. 
However, the percentage of questions that were assigned both the correct role and high grammaticality/adequacy were significantly higher for our model (around 10--20\% absolute).
As we will see in the error analysis below, these results follow in part from the baseline model overfitting to natural-sounding questions for explicit arguments (which are easier to make grammatical due to an abundance of training data), even when they are not appropriate for the role. 
We also find that adequacy takes a hit for implicit or None roles, as our model has seen few such examples during training and since often the placeholder-filler arguments are also implicit for those instances. 

\paragraph{Finding Implicit Arguments}
For explicit arguments in OntoNotes (69\% of questions), annotators selected a mismatched answer in 9\% of cases and marked the question unanswerable in 11\%.
For the non-explicit arguments, annotators quite often chose to answer the questions (50\% of cases).
36\% of these answers, which is 9\% of all questions, were deemed to be plausible implicit arguments via manual inspection. 
For example, consider the sentence ``It is only recently that the residents of Hsiachuotsu, unable to stand the smell, have begun to protest." Here annotators answered ``the smell" when asked \textit{Why did the residents of Hsiachuotsu protest?} (\pbrole{A1}). Such implicit annotations could thus conceivably increase annotation coverage by about 10\%, indicating that our Role Questions may be appealing as a way to help annotate implicit arguments.

\paragraph{Error analysis}
\label{erroranalysis}
To understand the models' role correspondence errors, we check for cases where each model produced identical questions for different roles of the same predicate.
64\% of predicate instances had at least one duplicate question under the baseline, as opposed to 6\% for our model.
Upon further examination, we found that the baseline systematically repeats questions for explicit arguments when prompted with a role whose argument is implicit or missing.
For example, for the predicate \predicate{give.01}, in a context where only \pbrole{A1} (thing given) was explicit (\textit{the position was given}), it predicted \textit{What was given?} for all core roles.
This shows that our prototype generation stage is essential for handling these phenomena.

While RC accuracy is good for both explicit and implicit Role Questions, QA accuracy is lower on ON5V and G\&C. On one hand, this may be due to the financial domain of G\&C and the fact that it targets nominal predicates, making it harder for annotators to understand. We also notice that the contextualizer sometimes fills the placeholders with the wrong argument from context, either because it is implicit or ambiguous. In such cases the annotators could mark the correct role, but do not answer the question properly.

In general, contextualization works well: The BART model is able to correctly identify tense, modality, negation and animacy in most cases. We inspected 50 randomly sampled instances of questions with average adequacy below 3, finding that the most common error 
is due to the placeholder being wrongly filled. Other errors are mainly due to an incorrect animacy judgment (\textit{who} vs. \textit{what}) or preposition or verb sense mistakes.

\paragraph{Modifiers}
We also evaluate results on 5 modifier roles for 100 predicate instances in OntoNotes.
On these, grammaticality (4.20), adequacy (4.29), and role accuracy (81\%) are comparable to results on core arguments, but QA accuracy (45\%) is much lower. However, this number is not very reliable: of 500 modifier role questions, <10\% corresponded to explicit arguments, because modifiers are relatively sparse in OntoNotes.

\subsection{Comparison to Related Systems}
\label{sec:comparisons}
To understand how our system fits in the landscape of existing work, we compare to two recently published question generation methods: Syn-QG \citep{dhole2020syn} and Event Extraction Questions \citep[EEQ]{du-cardie-2020-event}.
These comparisons require some care, as the systems differ from ours in scope and inputs/outputs: Syn-QG only generates questions for arguments detected in the text, and EEQ uses fixed, template-based phrases for roles in an ontology of event types (rather than broad-coverage semantic roles).

\paragraph{Comparison to Syn-QG}
Syn-QG \citep{dhole2020syn} uses several techniques, including off-the-shelf syntax, SRL, and NER models, to identify potential answers in a sentence and generate questions which ask about them (examples in \autoref{tab:exqcomparison}).

Reusing their published code, we validate the model's output with the authors\footnote{Following their advice, we exclude questions generated from the templates for WordNet supersenses, as they were a source of noise.} and apply it to a sample of 100 sentences from OntoNotes. We run their model on these sentences and collect 143 QA pairs where the answer in Syn-QG overlaps significantly with a gold SRL argument, and assign the gold role label to the paired question. Then we use our system to generate questions for these role labels and evaluate both sets of questions according to our metrics.

The system's output is evaluated using our evaluation criteria, where results are shown in \autoref{tab:dhole}.
Our model has better role accuracy (85\%) than Syn-QG (60\%), though this may be unsurprising since ours conditions on the gold role.
Perhaps more surprisingly, our model also strongly wins on QA accuracy, with 75\% to Syn-QG's 59\%,
despite Syn-QG conditioning directly on an answer that highly overlaps with the gold argument.
Furthermore, our Role Questions are deemed more grammatical and adequate on average.
These results suggest that our model has significant advantages over Syn-QG for fluently and correctly capturing aspects of semantic structure related to semantic roles.

\begin{table}[t]
\small
\resizebox{\columnwidth}{!}{%
\begin{tabular}{l|llll}
      & RA   & QA   & ADQ  & GRM  \\ \hline
RoleQ & 0.85 & 0.75 & 4.43 & 4.49 \\
SynQG & 0.60 & 0.59 & 3.57 & 4.19
\end{tabular}}
\caption{Comparison between our (RoleQ) approach and Syn-QG \citep{dhole2020syn} on 100 random frames in OntoNotes, covering 143 core arguments.
\textit{RA} is Role Accuracy, \textit{QA} is answer accuracy, \textit{ADQ} is adequacy, and \textit{GRM} is  grammaticality. }
\label{tab:dhole}
\end{table}

\paragraph{Comparison to EEQ}
\citet{du-cardie-2020-event} generate questions by applying fixed templates to argument descriptions in the ACE \citep{doddington2004automatic} event ontology, in order to facilitate the use of a QA system to extract arguments.
For example (\autoref{tab:exqcomparison}), the \textsc{Justice.Pardon} event has a \textsc{Defendant} argument which receives the question \textit{Who is the defendant?}
Event detection and extraction, in comparison to SRL, deals with more abstract events which may be expressed by a variety of predicates: for example, the \textsc{Personnel.Elect} event may be triggered by a phrase like ``came to power'' (\autoref{tab:exqcomparison}, second row), where the verbal predicate is the more general \predicate{come.04}.\footnote{In PropBank, \predicate{come.04} includes such expressions as \textit{come to fruition} and \textit{come under scrutiny}.}

For comparison, we randomly extracted two arguments from ACE for each event and role covered by their questions, for 198 total arguments.
Two of the authors then manually mapped the participant types to the corresponding PropBank roles of the predicate denoted by the annotated trigger word in ACE.\footnote{For example, in the context of ``coming to power,'' the \textsc{elected-entity} in ACE is mapped to \textsc{come.04-A0} in PropBank.}
We then evaluated both sets of questions according to our metrics, with the exception of role accuracy, since the EEQ questions are not specific to the trigger word.

Results are shown in \autoref{fig:Cardie}.
Our questions score higher in grammaticality and adequacy, as EEQ's template-based approach often results in awkward or ungrammatical questions like \textit{What declare bankruptcy?} (which, besides the subject agreement error, might need to ask \textit{who} in order to be adequate, depending on the context).
QA accuracy, on the other hand, is roughly comparable between the two systems for human annotators, showing that both do a similar job of capturing question semantics.
However, we also measure QA accuracy with an automated QA model trained on SQuAD 1.0 (QA SQuAD, \autoref{fig:Cardie}),
and we find that our contextualized questions produce much higher QA accuracy.
We suspect this is due to our contextualization step producing natural-sounding questions which are similar to those in other other QA datasets, aiding transfer.

\begin{table}[t]
    \centering
    \resizebox{\columnwidth}{!}{%
\begin{tabular}{l|lll|l}
      & GRM  & ADQ  & QA & QA SQuAD \\ \hline
RoleQ & 4.40 & 4.30 & 0.59         &     0.70                \\
EEQ   & 3.98 & 3.72 & 0.57         &     0.56               
\end{tabular}}
        \caption{Grammaticality, adequacy, and QA scores for our Role Questions (RoleQ) and EEQ \citep{du-cardie-2020-event}.
        We also report the QA Accuracy score of a SQuAD model. }
    \label{fig:Cardie}
\end{table}

\section{Conclusion}
We presented an approach to produce fluent natural language questions targeting any predicate and semantic role in the context of a passage. By leveraging the syntactic structure of QA-SRL questions in a two-stage approach, we overcome a lack of annotated data for implicit and missing arguments and produce questions which are highly specific to the desired roles.
This enables the automatic generation of information-seeking questions covering a large, broad-coverage set of semantic relations, which can bring the benefits of QA-based representations to traditional SRL and information extraction tasks.

\section*{Acknowledgments}
We would like to thank Daniela Brook-Weiss for helping in the initial stages of this project and the anonymous reviewers for their insightful comments. 
The work described herein was supported in part by grants from Intel Labs, Facebook, the Israel Science Foundation grant 1951/17.
This project has received funding from the Europoean Research Council (ERC) under the Europoean Union's Horizon 2020 research and innovation programme, grant agreements No. 802774 (iEXTRACT) and  No. 677352 (NLPRO).

\bibliographystyle{acl_natbib}
\bibliography{anthology,custom}

\appendix
\clearpage

\section{Aligning QA-SRL to SRL roles}\label{app:qasrl_role_align}
On the QA-SRL Bank 2.0, we aligned 190K QA-pairs with predicted SRL labels out of 271K gold questions. For QANom, we aligned 8.3K out of 21.5K gold questions. We use the verbal SRL parser in AllenNLP \cite{Gardner2018AllenNLPAD} which re-implements \citet{Shi2019SimpleBM}'s SRL parser. For nominal predicates, we re-train the same model on NomBank \cite{Meyers2004TheNP}, achieving 81.4 CoNLL-F1 score on the development set. 

\section{Converting QA-SRL Questions to Prototypes}
\label{sec:prototype_conversion}
To transform a QA-SRL question into its prototype,
we replace the AUX and VERB slot values with either \textit{is} and the past participle form (for passive voice), a blank and the present form (for active voice when SUBJ is blank), or \textit{does} and the stem form (for active voice when SUBJ is present). We also replace all occurences of \textit{who} and \textit{someone} with \textit{what} or \textit{something}.
This effectively removes all modality, aspect, negation, and animacy information, while converting all questions to the simple present tense. However, it preserves the active/passive distinction and other elements of the question's syntactic structure, which are relevant to the semantic role.

\section{Contextualizing QA-SRL Questions}
Here we provide extra details on the algorithm used to contextualize the questions provided by annotators in the QA-SRL Bank 2.0 and QANom.

\subsection{Resolving Syntactic Ambiguity}
\label{app:syntactic-ambiguity}
As written in \autoref{sec:contextualizing}, we first try to choose the syntactic structure that is shared with the greatest number of other questions.
If there is a tie (for example, if we only have one question for an instance), then we fall back to a few rules, depending on the type of ambiguity as follows:

\begin{itemize}
    \item \textbf{Preposition/Particle:} in a question like
    \textit{What does something give up?}, there is ambiguity over whether
    the object should be placed before or after the preposition (\textit{up}). Here we default to after the preposition, as any time the object position before the preposition is valid (e.g., \textit{something gives something up}), that means the preposition is acting as a particle, which generally admits the object after it as well (\textit{something gives up something}).
    \item \textbf{Locative arguments:} QA-SRL allows the placeholder \textit{somewhere} in the MISC slot. As a result, a question like \textit{Where does something put something?} is ambiguous between treating \textit{where} as denoting an adverbial (which would lead to the clause \textit{Someone put something}) or a locative argument (which would result in \textit{Someone put something somewhere}). We default to the adverbial interpretation.
    \item \textbf{Ditransitives:} the last type of ambiguity is for questions over ditransitive verbs like \textit{What did someone give someone?} which are ambiguous over which of the two objects is extracted, \ie, whether it should be \textit{Someone gave something someone} or (\textit{Someone gave someone something}).
    We default to resolving \textit{who} questions to the first object and \textit{what} questions to the second, to match English's tendency to put (generally animate) recipients/benefactors in the first object position and (generally inanimate) themes in the second.
\end{itemize}

\subsection{Aligning Answers to Placeholders}
\label{app:aligning-answers}
After resolving ambiguities,
every placeholder position in a QA-SRL question and every answer to a QA-SRL question is associated with a syntactic function (\role{subj} for the SUBJ slot, \role{obj} for the OBJ slot, or \role{pp}/\role{xcomp}/\role{obj2}/\role{loc} for the PREP/MISC slots --- see \autoref{tab:qasrl}) and syntactic structure (which we may denote by the question's declarative form).
To produce Frame-Aligned QA-SRL questions, we replace the placeholders in each question with any answers bearing the same syntactic function in the same syntactic structure (see \autoref{fig:aligned-qasrl-example}).
For the purpose of encoding these syntactic structures, we ignore the same factors as we strip out in the question prototyping step (\autoref{sec:generating_protos}): tense, modality, negation, and animacy (but we keep the active/passive distinction).

To further increase coverage of our placeholder alignments, we add extra correspondences between syntactic structures which correspond to the same semantic role in many cases:
\begin{itemize}
    \item We align the \role{obj} of a transitive clause (with a \role{loc} or no MISC argument) with the \role{subj} of passive clauses (with a \role{loc}, \textit{by}-\role{pp}, or no MISC argument).
    \item We align the \role{subj} of a transitive clause (no MISC argument) with the prepositional object of a \textit{by}-\role{pp} in a passive clause.
    \item We align the \role{loc} argument of a transitive clause with the \textit{where}-adverbial of a transitive clause with no MISC.
    \item Finally, we align any \role{subj} arguments as long as their syntactic structures agree after stripping the PREP/MISC argument from both (and similarly for \role{obj}).
    For example, if we have \textit{Who brought something? / John}, then \textit{Who did someone bring something to?} would align with the answer to produce \textit{Who did John bring something to?}
\end{itemize}
Without these extra correspondences, our method populates 83.7\% of placeholders in the QA-SRL Bank 2.0 and QANom, while with them, we cover 91.8\%.

\paragraph{Grammar Correction}
We add two extra postprocessing steps to improve the quality of the questions.
First, before substituting answers for placeholders, we attempt to undo sentence-initial capitalization by decapitalizing the first word of sentences in the source text if both the second character of the first word is lowercase (ruling out acronyms) and the first character of the second word is lowercase (ruling out many proper nouns).
Second, we fix subject/verb agreement for questions with plural subjects (as QA-SRL questions always have singular agreement)
by masking out the auxiliary verb (e.g., \textit{does}) and replacing it with the form that receives higher probability under a masked language model (e.g., either \textit{do} or \textit{does}).

\subsection{Training the Contextualizer}
\label{app:training-contextualizer}
We fine-tune BART on the Frame-Aligned QA-SRL dataset for 3 epochs on 4 GeForce GTX 1080Ti GPUs with an effective batch size of 32 and maximum target sequence length of 20. We use the standard separator token between the question and the passage. We also surround the predicate token with text markers \textit{PREDICATE-START} and \textit{PREDICATE-END} (but without using new embeddings in the vocabulary), and insert the predicate lemma again right after the question. The full text input may look like:
Some geologists PREDICATE-START study PREDICATE-END the Moon . \textlangle{}/s\textrangle{} study [SEP] what studies something ?, with the output question \textit{Who studies the moon?}.

\section{Annotation Interfaces}
\label{app:interfaces}

\begin{figure}[!htb]
\includegraphics[width=3in]{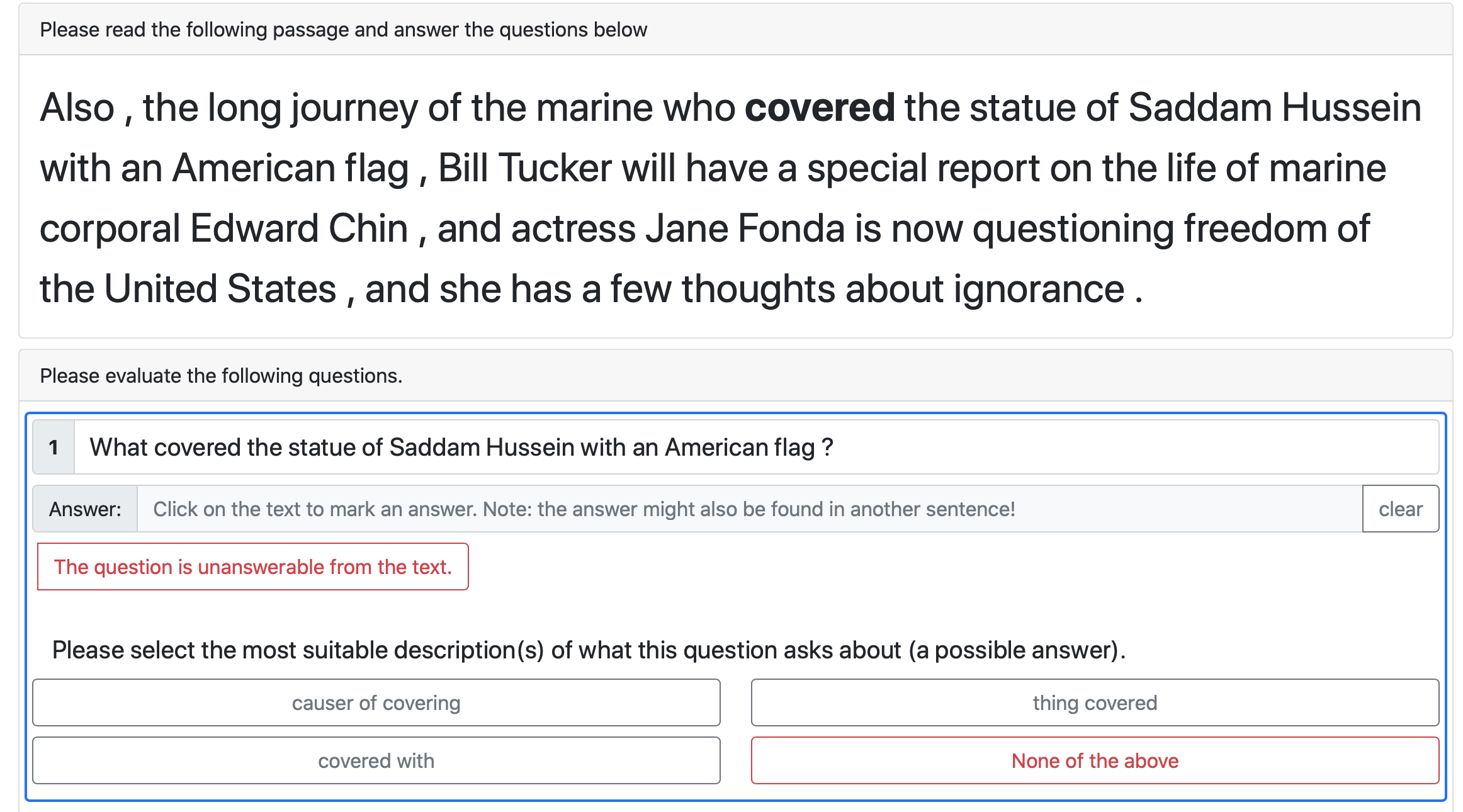}
\caption{Interface for QA and RC annotation.}
\label{fig:qa-rc-interface}
\end{figure}

\begin{figure}[!htb]
\includegraphics[width=3in]{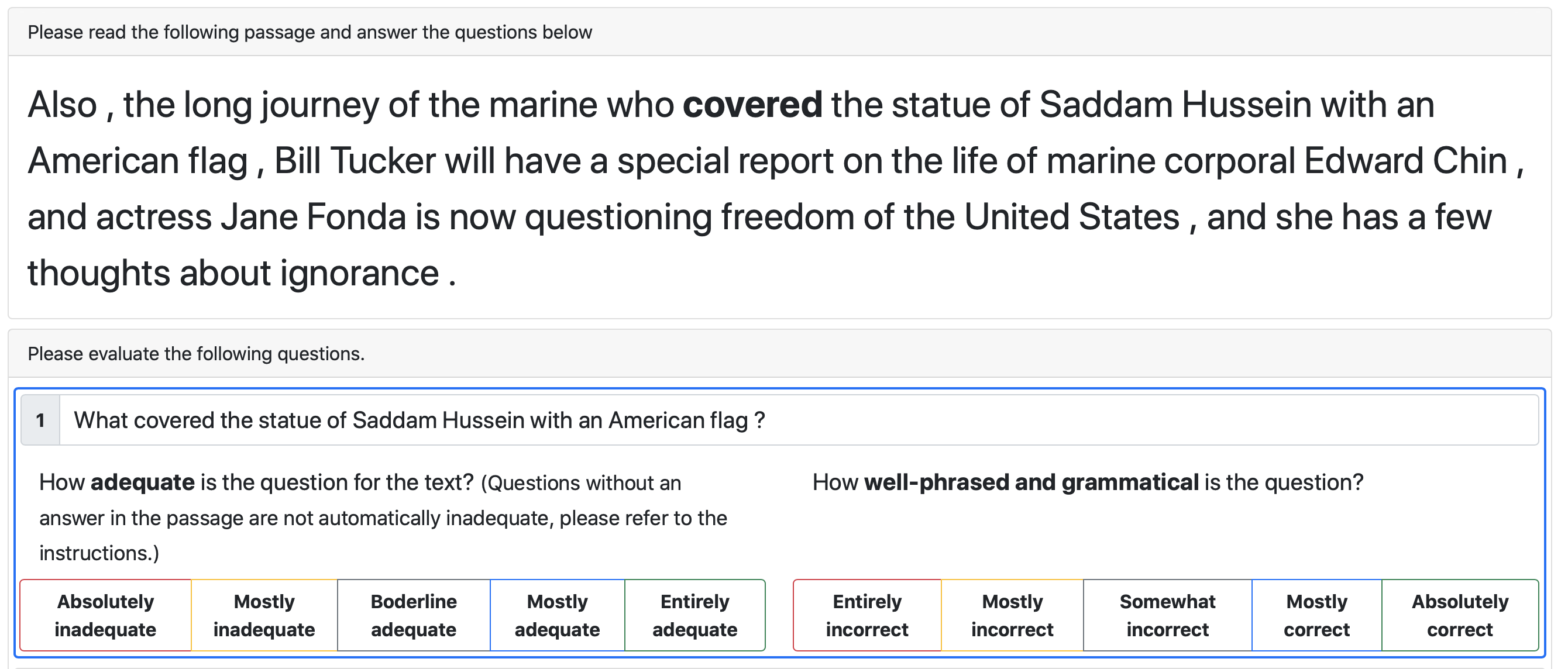}
\caption{Interface for grammaticality and adequacy annotation.}
\label{fig:grm-adq-interface}
\end{figure}

\section{Coverage}
OntoNotes includes 530K core argument instances with 11K distinct roles. Our question lexicon contains prototypical questions for almost all arguments except 4K ($<$1\%) instances. When filtering out questions with less than 50\% SQuAD-F1 accuracy, the leftover prototypical questions cover 83.5\% of all argument instances.

\end{document}